%% file: jmlr-sample.tex
 \documentclass[pmlr,twocolumn,10pt,x11names,table]{jmlr} 

\usepackage{booktabs}
\usepackage{hyperref}
\usepackage{multirow}

\usepackage{siunitx}
\usepackage{subcaption}
\usepackage[switch]{lineno}

\usepackage{float}

\newcommand{\equal}[1]{{\hypersetup{linkcolor=black}\thanks{#1}}}

\theorembodyfont{\upshape}
\theoremheaderfont{\scshape}
\theorempostheader{:}
\theoremsep{\newline}

\jmlrvolume{297}
\jmlryear{2025}
\jmlrworkshop{Machine Learning for Health (ML4H) 2025} 

 \title[FeatureEndo-4DGS]{FeatureEndo-4DGS: Real-Time Deformable Surgical
Scene Reconstruction and Segmentation with 4D
Gaussian Splatting}

\author{%
\Name{Kai Li}\equal{These authors contributed equally} \Email{kaii.li@mail.utoronto.ca}\\
\addr University of Toronto, Canada
\AND
\Name{Junhao Wang}\footnotemark[1] \Email{hry.wang@mail.utoronto.ca}\\
\addr University of Toronto, Canada
\AND
\Name{William Han} \Email{wjhan@andrew.cmu.edu}\\
\addr Carnegie Mellon University, USA
\AND
\Name{Ding Zhao} \Email{dingzhao@cmu.edu}\\
\addr Carnegie Mellon University, USA
}

\begin{document}

\maketitle

\begin{abstract}
Minimally invasive surgery (MIS) requires high-fidelity, real-time visual feedback of dynamic and low-texture surgical scenes. To address these requirements, we introduce FeatureEndo-4DGS (FE-4DGS), the first real-time pipeline leveraging feature-distilled 4D Gaussian Splatting for simultaneous reconstruction and semantic segmentation of deformable surgical environments. Unlike prior feature-distilled methods restricted to static scenes, and existing 4D approaches that lack semantic integration, FE-4DGS seamlessly leverages pre-trained 2D semantic embeddings to produce a unified 4D representation—where semantics also deform with tissue motion. This unified approach enables the generation of real-time RGB and semantic outputs through a single, parallelized rasterization process. Despite the additional complexity from feature distillation, FE-4DGS sustains real-time rendering \textbf{(287.95 FPS)} with a compact footprint, achieves state-of-the-art rendering fidelity on EndoNeRF \textbf{(39.1 PSNR)} and SCARED \textbf{(27.3 PSNR)}, and delivers competitive EndoVis18 segmentation, matching or exceeding strong 2D baselines for binary segmentation tasks \textbf{(0.93 DSC)} and remaining competitive for multi-label segmentation \textbf{(0.77 DSC)}. 
\end{abstract}
\begin{keywords}
Surgical Scene Reconstruction, Feature Distillation, Gaussian Splatting
\end{keywords}


\paragraph*{Data and Code Availability}
This study did not involve the collection of new data from human subjects; all data is obtained from publicly available datasets. Specifically, we used the EndoNeRF dataset for endoscopic 4D reconstruction \citep{wang2022neuralrenderingstereo3d}, the SCARED dataset for robotic surgery depth estimation \citep{allan20202018roboticscenesegmentation}, and the EndoVis18 dataset for surgical image segmentation \citep{allan20202018roboticscenesegmentation}. Our code for reproducing all experiments, including pre-processing and model training will be available at the following link: \url{https://github.com/kailathan/FE-4DGS.}


\paragraph*{Institutional Review Board (IRB)}
This research did not involve the collection of new data from human subjects. All data used in this study were obtained from publicly available, de-identified datasets, and therefore IRB approval was not required.

\section{Introduction}
\label{sec:intro}
Recent advancements in artificial intelligence (AI) and the decreasing cost of computational resources have enabled the development of automated methods to assist minimally invasive surgery (MIS) in real-time \citep{ali2023comprehensivesurveyrecentdeep}. Tasks such as image classification \citep{subedi2024classificationendoscopyvideocapsule}, object detection \citep{yu2022colonoscopypolypdetectionmassive}, semantic segmentation \citep{zhu2024medicalsam2segment}, tissue tracking \citep{wang2024endogslamrealtimedensereconstruction}, and surgical scene reconstruction \citep{zha2023endosurfneuralsurfacereconstruction} have become increasingly sophisticated, significantly benefiting robotic-assisted MIS.

Surgical scene reconstruction has recently attracted considerable attention \citep{zha2023endosurfneuralsurfacereconstruction, long2021edssrefficientdynamicsurgical, liu2020reconstructingsinusanatomyendoscopic, liu2024endogaussianrealtimegaussiansplatting}. This technique provides surgeons with a comprehensive real-time view that enhances navigation and instrument control, and has potential to enable robotic surgery automation. Early methods were dominated by traditional techniques such as simultaneous localization and mapping (SLAM) \citep{CHEN2018135, Song_2018, 10.1007/978-3-030-87202-1_32}, but the field has since transitioned to neural network–based approaches \citep{long2021edssrefficientdynamicsurgical, Li_2020} and more recently, to neural radiance fields (NeRFs) \citep{mildenhall2020nerfrepresentingscenesneural, zha2023endosurfneuralsurfacereconstruction}. However, NeRF-based methods require large volumes of data and suffer from slow rendering speeds, motivating the development of 3D Gaussian Splatting (3DGS) \citep{kerbl20233dgaussiansplattingrealtime} as a more efficient alternative. Landmark studies such as EndoGaussian \citep{liu2024endogaussianrealtimegaussiansplatting} and LGS \citep{10.1007/978-3-031-72384-1_62} have successfully applied 3DGS to surgical scene reconstruction.

Feature field distillation techniques, which were initially developed for the NeRF domain \citep{zhi2021inplacescenelabellingunderstanding, siddiqui2022panopticlifting3dscene, kobayashi2022decomposingnerfeditingfeature}, have been extended to 3DGS \citep{10.1007/978-3-031-72658-3_19, zhou2024feature3dgssupercharging3d} beyond their application in surgical scene reconstruction. Inspired by these advances, we propose FeatureEndo-4DGS (FE-4DGS). FE-4DGS integrates semantic features of 2D foundation models \citep{ravi2024sam2segmentimages, zhu2024medicalsam2segment} into 4DGS to support real-time segmentation and rendering of deformable surgical scenes. Crucially, online segmentation enables safe surgical augmented reality (AR) renderings by highlighting critical anatomy and instruments to guide surgical actions \citep{Doornbos2024ARMIS}. We introduce a unified deformation module to simultaneously update both per-Gaussian geometric properties including position, scale, rotation, and opacity, in addition to per-Gaussian semantic features. The module extracts deformation features, which are learned deformation cues that enable semantic features and geometry to be updated consistently and efficiently. During optimization, we further employ a CNN-based semantic decoder to align the rendered semantic feature map with features extracted by 2D segmentation models.

In summary, our contributions are as follows:\\
1. We develop FE-4DGS, a real-time pipeline that jointly reconstructs deformable scenes and semantic feature fields in MIS in a single pass. \\
2. We achieve state-of-the-art performances on rendering fidelity, while maintaining real-time rendering speeds.\\
3. We compare FE-4DGS and 2D segmentation foundation models on binary and multi-label segmentation tasks and achieve state-of-the-art performances on binary segmentation, while being competitive in multi-label segmentation.

\section{Related Works}
\subsection{Surgical Scene Reconstruction}
Reconstructing soft tissues from endoscopic stereo videos is critical for intra\-operative navigation and robotic automation~\citep{lu2021superdeepsurgicalperception, 10.3389/frobt.2017.00015, liu2020reconstructingsinusanatomyendoscopic}. Traditional approaches based on SLAM~\citep{CHEN2018135, Song_2018, 10.1007/978-3-030-87202-1_32, gao2019surfelwarpefficientnonvolumetricsingle} degrade under non-rigid motion, specular reflections, and occlusions. Learning-based methods with CNNs and vision transformers improve stereo depth estimation and non-rigid modeling~\citep{Li_2020}, yet remain limited in dynamic, unstructured scenes with weak depth cues.

NeRFs~\citep{mildenhall2020nerfrepresentingscenesneural} offer powerful 3D representations of complex deformations and have been extended to dynamic objects in nonsurgical contexts~\citep{niemeyer2021girafferepresentingscenescompositional, martinbrualla2021nerfwildneuralradiance}. EndoNeRF~\citep{wang2022neuralrenderingstereo3d} adapts dynamic NeRFs for robotic surgery, achieving state-of-the-art 3D reconstruction and deformation tracking. However, NeRFs demand extensive point and ray sampling, leading to heavy computational costs that hinder real-time applications~\citep{chen2024survey3dgaussiansplatting, rabby2024beyondpixelscomprehensivereviewevolution}.

To address efficiency constraints, 3DGS~\citep{kerbl20233dgaussiansplattingrealtime} models scenes with anisotropic Gaussians and tile-based rasterization, enabling real-time rendering while preserving quality. Under deformable conditions, recent works have leveraged 4DGS for surgical reconstruction, demonstrating strong performance in real-time 3D video scene modeling~\citep{xie2024surgicalgaussiandeformable3dgaussians, chen2024surgicalgsdynamic3dgaussian}.

\subsection{Surgical Scene Segmentation}
In addition to real-time scene reconstruction, simultaneous segmentation of key anatomical structures and surgical instruments during intraoperative procedures is highly advantageous. Previous works have explored segmentation in endoscopic surgeries \citep{10.1007/978-3-031-43996-4_10, 10.1007/978-3-031-72114-4_12, CHEN2024103310}. With recent advances in segmentation foundation models, such as SAM \citep{kirillov2023segment} and SAM 2 \citep{ravi2024sam2segmentimages}, and their adaptations to the medical domain (e.g., MedSAM \citep{Ma_2024}, MedSAM 2 \citep{zhu2024medicalsam2segment}), many approaches have fine-tuned these models for surgical applications. A recent study, Feature 3DGS \citep{zhou2024feature3dgssupercharging3d}, extends 3DGS to incorporate feature field distillation, enabling real-time segmentation, language-guided editing, and other interactive operations. Building on these developments, our work integrates 4DGS (i.e., deformable 3DGS) with feature field distillation to facilitate real-time rendering and segmentation of surgical scenes, thereby advancing intraoperative visualization.

\begin{figure*}[!t]
    \centering
    \includegraphics[width=1\linewidth]{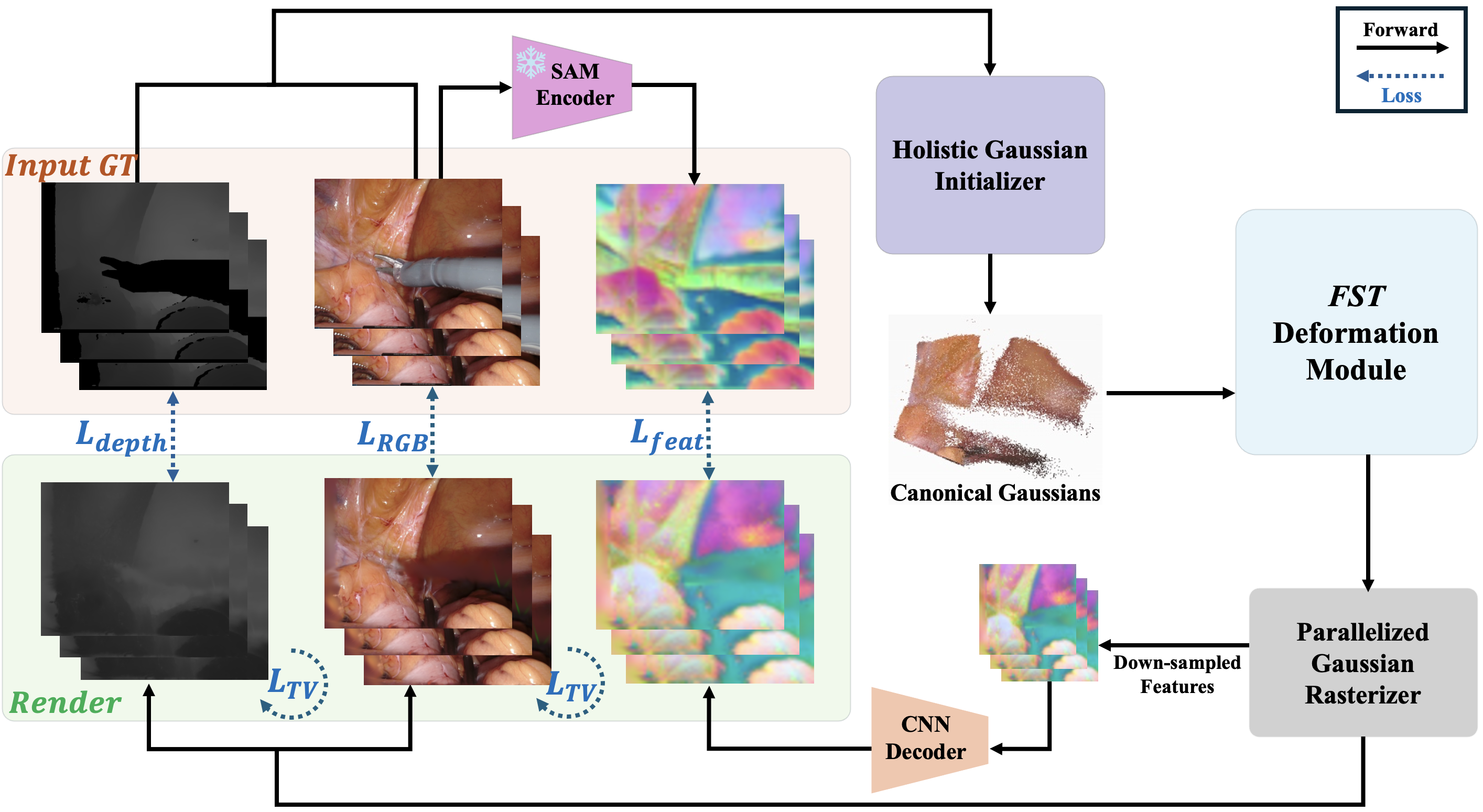}
    \caption{Overview of FE-4DGS. The pipeline begins with holistic Gaussian initialization, re-projecting image pixels into 3D Gaussians. In the FST deformation module, a 4D voxel encoder extracts latent features, which a deformation decoder refines by updating Gaussian parameters and semantics via a lightweight MLP (Section~\ref{sec:fst}, Appendix~\ref{apd:fst}). A differentiable rasterizer renders the updated Gaussians into radiance and semantic maps, which a CNN decoder upsamples and aligns with features from a 2D segmentation model (SAM) to ensure semantic consistency.}
    \label{fig:feg}
\end{figure*}

\section{Our Method}
We propose FE-4DGS, a 4D Gaussian splatting framework that augments surgical scene reconstruction with dense, real-time semantic features distilled from 2D segmentation foundation models (denoted as SAM for simplicity). Our pipeline integrates semantics through two components: (1) a Gaussian deformation module with a motion-aware decoder \(F_{\text{feat}}\) that updates per-Gaussian features across frames, and (2) a CNN-based decoder that upsamples rendered semantic maps and aligns them with SAM outputs under a per-pixel \(L_1\) loss. A parallelized rasterizer jointly renders color, depth, and semantic features, enabling comprehensive scene representation. An overview of the pipeline is shown in Figure~\ref{fig:feg}.

\subsection{3D Gaussian Scene Representation and Initialization}
Our method builds upon the 3D Gaussian Splatting (3DGS) framework \citep{kerbl20233dgaussiansplattingrealtime}, which represents a scene using a dense collection of 3D Gaussians. Each Gaussian is centered at a mean \(\mu\) and characterized by a covariance matrix \(\Sigma\) that defines its spatial spread. Specifically, the contribution of a Gaussian at a 3D point \(x\) is given by:
\begin{equation}
  G(\mathbf{x}) = \exp\!\Bigl(-\frac{1}{2} (x\ - \mu)^\top \Sigma^{-1} (x\ - \mu)\Bigr).
\end{equation}
For both efficiency and interpretability, the covariance matrix is factorized as:
\begin{equation}
  \Sigma = R\,S\,S^\top\,R^\top,
\end{equation}
where \(R\) is a rotation matrix and \(S\) is a scaling matrix. In this representation, the scene is modeled by jointly optimizing the parameters of each Gaussian—its position \(\mu\), rotation \(R\), scaling \(S\), opacity \(o\), and spherical harmonic (SH) coefficients.

Following \cite{liu2024endogaussianrealtimegaussiansplatting}, a holistic initialization is performed by re-projecting pixels from the input image sequence using estimated depth maps. For an image \(I_i\) with depth \(D_i\) and binary mask \(M_i\), the corresponding 3D points are computed as:
\begin{equation}
  P_i = K^{-1}T_i\,D_i\,\bigl(I_i \odot M_i\bigr), \quad P = \bigcup_{i=1}^{T} P_i,
\end{equation}
where \(K\) and \(T_i\) denote the intrinsic and extrinsic parameters of the camera, respectively.

\subsection{Feature-Spatiotemporal (FST) Deformation Module: Semantic Feature Integrated Deformation Decoder}
\label{sec:fst}
To capture spatiotemporal deformation, each Gaussian is first encoded by a 4D voxel module that maps its center \(\mu\) and time \(t\) to a latent feature \(f\). 
Inspired by \citet{yang2023neurallerplanerepresentationsfast, wu20244dgaussiansplattingrealtime} to represent the 4D voxel module as a multi‐resolution HexPlane introduced by \citet{cao2023hexplanefastrepresentationdynamic}, we leverage a multi-resolution HexPlane to efficiently learn features for Gaussians in spacetime. The HexPlane is paired with a light MLP decoder to produce a hidden representation \(h = F_{\mathrm{out}}(f)\in\mathbb{R}^W\) that is fed into smaller deformation branches. Here, W denotes size of the hidden dimension. Refer to Appendix~\ref{apd:fst} for more details.

\smallskip
\noindent\textbf{Four-Branch Geometric Deformation Decoder:} 
We utilize four lightweight decoder branches to update the position $\mu$, rotation $R$, scaling $S$, and opacity $o$. Let 
$h = F_{\mathrm{out}}(f) \in \mathbb{R}^W$ denote the intermediate feature representation. For each geometric property $g \in \{\mu, R, S, o\}$, a branch-specific feature extractor $F_g^{\mathrm{feat}}$ and prediction head $F_g^{\mathrm{head}}$ produce the corresponding update $\Delta g$:

\begin{align}
h_g &= F_g^{\mathrm{feat}}(h), & 
\Delta g &= F_g^{\mathrm{head}}(h_g) \in \mathbb{R}^{d_g}, \\ \nonumber
\end{align}

\noindent where the output dimensions are $d_\mu = 3$, $d_R = 4$, $d_S = 3$, and $d_o = 1$. The features $h_g$ are also used to generate semantic updates, as described in the next section.

\smallskip
\noindent\textbf{Feature‐Based Semantic Updater:}
Within a 4D dynamic surgical scene, semantic embeddings derived from SAM undergo systematic shifts under changes in viewpoint and 3D deformation throughout time. To address this, we introduce a semantic update network \(F_{\mathrm{feat}}\) which utilizes the aforementioned deformation features. We concatenate:
\[
  u_{\mathrm{all}}
  = \bigl[h_\mu \,\|\, h_R \,\|\, h_S \,\|\, h_o\bigr]
  \;\in\;\mathbb R^{4W},
\]
\smallskip
which is passed through our semantic‐update network \(F_{\mathrm{feat}}\) to obtain the per-Gaussian semantic feature update \({\Delta{z}}\in\mathbb{R}^N\) where $N$ is the number of feature channels:
\begin{equation}
\label{eq:5}
  \Delta z 
  = F_{\mathrm{feat}}\bigl(u_{\mathrm{all}})\in\mathbb{R}^N\,
  \quad
  z' = z + \Delta z.
\end{equation}

Finally, the fully augmented Gaussians are updated as follows:
\smallskip
\begin{equation}
  G_t = \bigl(\mu + \Delta\mu,\;
              R + \Delta R,\;
              S + \Delta S,\;
              o + \Delta o,\;
              z'\bigr).
\end{equation}

\noindent\textbf{Semantic Feature Extraction:} For all experiments, we used a pre-trained SAM~\citep{kirillov2023segment} with a ViT-H encoder to extract semantic feature maps. Given an input image, SAM produces a high-dimensional semantic feature map with a spatial resolution of $64 \times 64$ with 256 channels. To match the aspect ratio of the input images, we adopt the cropping procedure from~\citet{zhou2024feature3dgssupercharging3d}: an input image of size $H \times W$ (with $W > H$) yields a cropped feature map of size $64 \times \frac{64W}{H}$, preserving semantic information without introducing padding artifacts. The resulting feature maps serve as supervision signals during training. To improve computational efficiency without sacrificing semantic expressivity, we compress the rendered feature maps to 128 dimensions, which are updated through the semantic MLP branch $F_{\text{feat}}$ as defined in Equation~\ref{eq:5}.

\subsection{Differentiable Rendering and Loss Functions}
During rendering, both the radiance (color) and semantic feature fields are computed via front-to-back alpha blending. For a pixel \(x\), the rendered color \(\hat{C}(x\)) and semantic feature \(\hat{z}(x\)) are given by:
\begin{align}
  \hat{C}(x) &= \sum_{i=1}^{N} c_i\,\alpha_i \prod_{j=1}^{i-1} \bigl(1-\alpha_j\bigr), \\
  \hat{z}(x) &= \sum_{i=1}^{N} z'_i\,\alpha_i \prod_{j=1}^{i-1} \bigl(1-\alpha_j\bigr),
\end{align}
where \(c_i\) and \(z'_i\) denote the color and updated semantic feature contributions from the \(i\)th Gaussian, and \(\alpha_i\) is its effective opacity. The effective opacity is computed by evaluating a corresponding 2D covariance matrix:
\begin{equation}
  \Sigma'_i = J\,W\,\Sigma_i\,W^\top\,J^\top,
\end{equation}
with \(J\) representing the Jacobian of the affine approximation of the projective transformation and \(W\) the view transformation matrix.

\smallskip
\noindent\textbf{CNN-Based Semantic Decoder:} After differentiable rasterization, a CNN-based decoder is employed to upsample the rendered semantic feature map, matching the channel dimension to that of SAM. This decoder performs a simple pointwise convolution aligning the feature dimensionality to match SAM's output. The resulting semantic feature map is then compared, using a per-pixel \(L_1\) loss, to the high-level semantic features \(f_{\text{SAM}}\):
\begin{equation}
  \mathcal{L}_{\text{feat}} = \frac{1}{HW} \sum_{x \in \Omega} \bigl\|\hat{z}(x) - z_{\text{SAM}}(x)\bigr\|_1,
\end{equation}
where \(\Omega\) denotes the set of pixel coordinates in an image of resolution \(H\times W\).

In addition to the photometric loss \(\mathcal{L}_{\text{rgb}}\) and depth loss \(\mathcal{L}_{\text{depth}}\), as in~\citet{liu2024endogaussianrealtimegaussiansplatting}, the overall loss is defined as:
\begin{equation}\label{eq:loss}
\begin{split}
\mathcal{L} &= \lambda_{\text{rgb}}\,\mathcal{L}_{\text{rgb}}
+ \lambda_{\text{depth}}\,\mathcal{L}_{\text{depth}} \\
&\qquad + \lambda_{\text{feat}}\,\mathcal{L}_{\text{feat}}
+ \lambda_{\text{TV}}\,\mathcal{L}_{\text{TV}}.
\end{split}
\end{equation}
with \(\mathcal{L}_{\text{TV}}\) enforcing spatiotemporal smoothness. The hyperparameters are fixed at $\lambda_{\text{rgb}}=1$, $\lambda_{\text{depth}}=0.01$, $\lambda_{\text{feat}}=1$, and $\lambda_{\text{TV}}=0.03$.

\subsection{Training and Inference}

The complete optimization is performed in a coarse-to-fine manner. In the coarse stage, only the basic Gaussian parameters are updated. In the fine stage, the semantic branch—comprising both the MLP \(F_{\text{feat}}\) within the deformation decoder and the CNN-based semantic decoder—is activated. This enables the network to jointly optimize the 3D Gaussian parameters and all MLP weights (including those of \(F_{\text{feat}}\)) using the Adam optimizer, while the CNN-based decoder ensures that the rendered semantic feature map is accurately aligned with the SAM features.

During inference, segmentation masks are generated by decoding the rendered and upsampled semantic features using SAM's pretrained decoder. To facilitate accurate multi-instance segmentation, class-specific bounding box prompts derived from ground truth annotations are provided as input. The inference process can be summarized as:
\begin{align}
\hat{z}(x) &= \text{FE-4DGS}_{\text{render}}(G_t, x), \\
\hat{z}'(x) &= \text{FE-4DGS}_{\text{decoder}}(\hat{z}(x)), \\
\hat{M}(x) &= \text{SAM}_{\text{decoder}}(\hat{z}'(x), B),
\end{align}
where $\hat{z}(x)$ represents the rendered semantic feature at pixel $x$, $\hat{z}'(x)$ is the upsampled feature map, $B$ denotes bounding box prompts, and $\hat{M}(x)$ is the resulting segmentation mask.

We follow the experimental setup and configurations from EndoGaussian~\citep{liu2024endogaussianrealtimegaussiansplatting} for evaluation consistency, noting that further hyperparameter tuning on specific datasets, such as the EndoVis18 dataset \citep{allan20202018roboticscenesegmentation}, may yield additional performance improvements.

\section{Experiments}
\label{sec:exp}

\subsection{Experimental Setup}
We optimize both the original Gaussian features and the deformation module—including parameters for HexPlane \citep{cao2023hexplanefastrepresentationdynamic}, feature extractors, and decoders—using the Adam optimizer \citep{kingma2017adammethodstochasticoptimization}. Following \cite{liu2024endogaussianrealtimegaussiansplatting}, we first optimize the canonical (coarse) Gaussians for 1000 iterations and then train the full FE-4DGS model for an additional 6000 iterations. In each iteration, we render one training camera view, compute losses, and apply an optimizer step. To ensure stable convergence, we apply an exponential learning rate decay schedule. All experiments are conducted on NVIDIA RTX A6000 48GB and NVIDIA RTX 5090 32GB GPUs unless specified otherwise.
We provide a detailed breakdown of all hyperparameters used, including learning rates for different parameter groups, decay schedules, and other tuning adjustments in Appendix~\ref{apd:hp}.

\input{tables/main_1}
\input{tables/main_2}

\begin{figure*}[ht]
    \centering
    \includegraphics[width=1\linewidth]{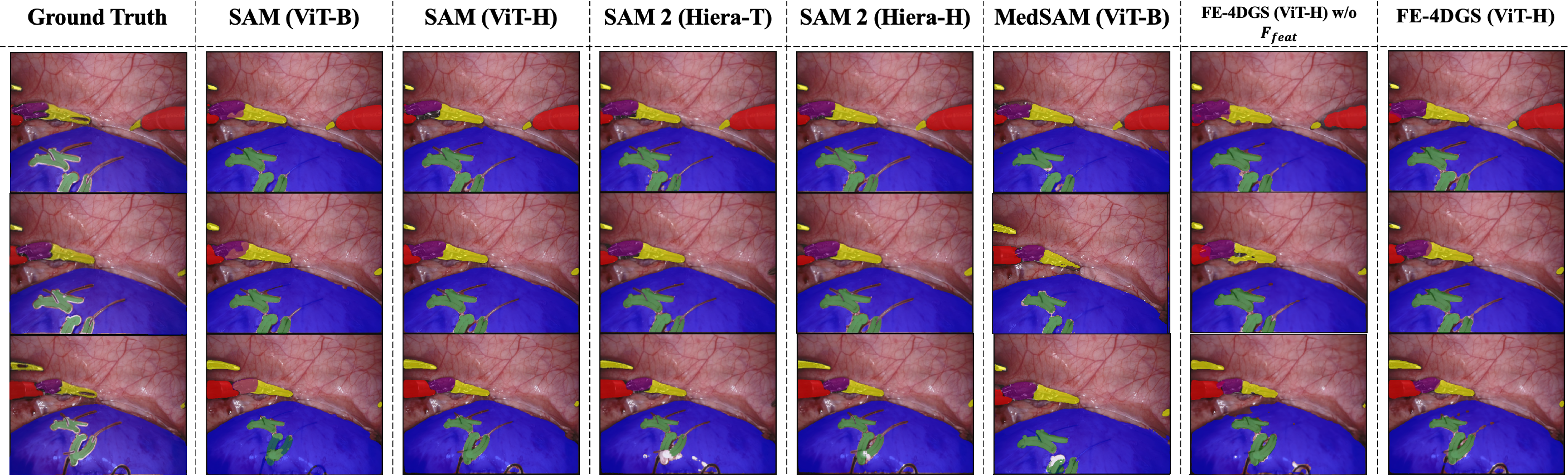}
    \caption{Qualitative segmentation comparisons. Colors correspond to the following classes: kidney (blue), small intestine (orange), instrument shaft (red), instrument clasper (yellow), instrument wrist (purple), and clamps (green). Notably, FE-4DGS (ViT-H) and SAM (ViT-H) exhibit clean multi-label segmentations, while other models struggle with finer labels such as clamps or instrument claspers. }
    \label{fig:qual}
\end{figure*}

\input{tables/main_3}
\input{tables/combined2}

\subsection{Datasets}
For surgical scene reconstruction, we use the EndoNeRF \citep{wang2022neuralrenderingstereo3d} and SCARED \citep{allan2021stereocorrespondencereconstructionendoscopic}. For segmentation tasks, we utilize the EndoVis18 \citep{allan20202018roboticscenesegmentation}.

We utilize the cutting and pulling sets from EndoNeRF, which consist of 2 in-vivo prostatectomy cases captured with stereo cameras from a single viewpoint. The pulling and cutting sets contain 63 and 156 frames, respectively. Following \citet{zha2023endosurfneuralsurfacereconstruction}, we select every eighth frame for testing.

The SCARED dataset comprises RGB-D scans of porcine cadaver abdominal anatomy and includes 7 sequences. In line with protocols from EndoGaussian \citep{liu2024endogaussianrealtimegaussiansplatting} and LGS \citep{10.1007/978-3-031-72384-1_62}, we use a subset of 5 sequences—datasets 1, 2, 3, 6, and 7. For each selected sequence, we render the first keyframe and use all frames within that keyframe, denoted as d1k1 (197 frames), d2k1 (88 frames), d3k1 (329 frames), d6k1 (637 frames), and d7k1 (647 frames). A 7:1 train-test split is adopted, consistent with prior works (i.e., \citet{liu2024endogaussianrealtimegaussiansplatting}).

For segmentation evaluation, we use the EndoVis18 dataset, which contains 4 sets of 149 frames of diverse surgical scenes. Sets 2-4 are used for training and set 1 is reserved for testing, including a 60-frame video to assess our integrated rendering and segmentation pipeline. Since EndoVis18 is designed for segmentation rather than scene reconstruction, it lacks per-frame camera calibrations and poses; motion blur and low frame rates further complicate reconstruction. We therefore report metrics between ground-truth semantic masks and decoded semantic features rendered by FE-4DGS during training, and compare them against the zero-shot performance of 2D segmentation foundation models. 2 tasks are considered: multi-label segmentation with 6 classes (kidney, small intestine, instrument shaft, instrument clasper, instrument wrist, clamps) and binary segmentation with all foreground classes merged.

\begin{figure*}[!ht]
    \centering
    \includegraphics[width=0.95\linewidth]{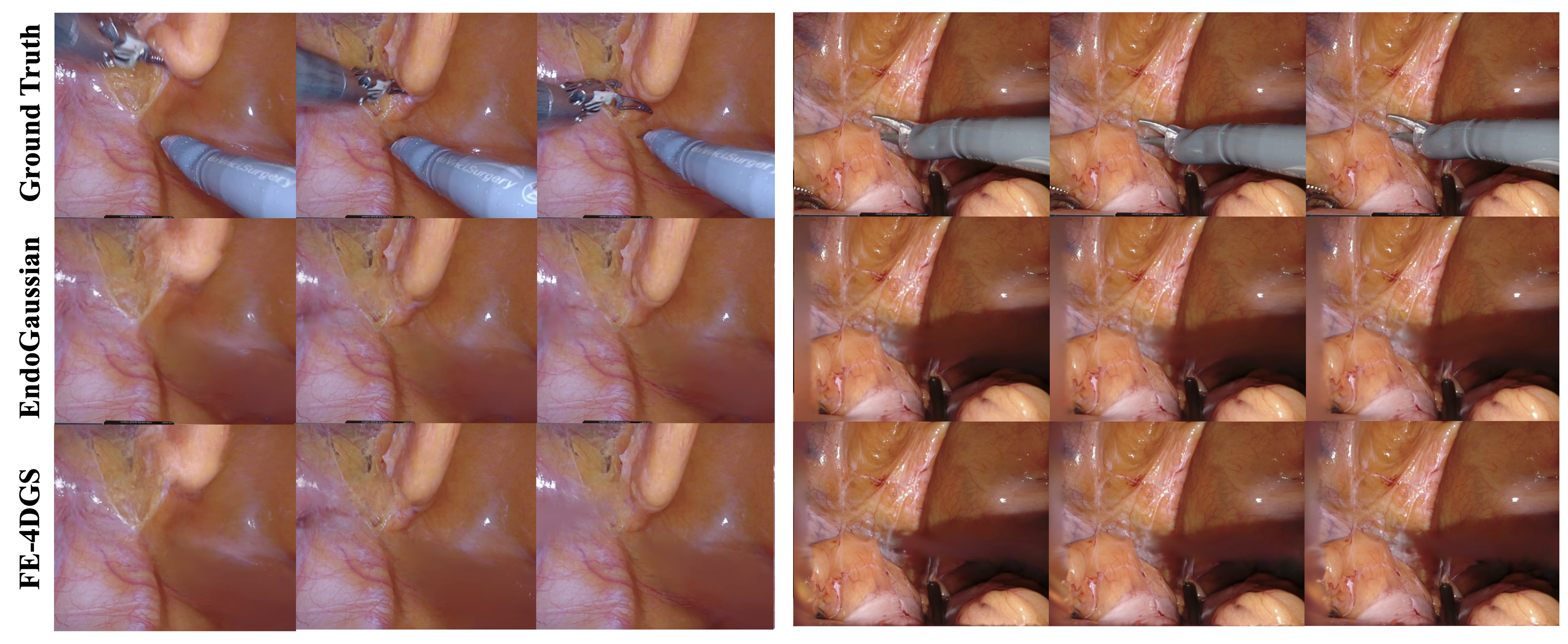}
    \caption{Comparison of qualitative renderings between EndoGaussian \citep{liu2024endogaussianrealtimegaussiansplatting}, FE-4DGS, and ground truth on the cutting (right) and pulling (left) sets of the EndoNeRF dataset \citep{wang2022neuralrenderingstereo3d}. Compared to previous methods, specular regions are reconstructed more finely in FE-4DGS. }
    \label{fig:render}
\end{figure*}

\subsection{Main Results}
We now discuss the performance of FE-4DGS in both rendering and segmentation, comparing our method against existing approaches and evaluating various ablations.
\paragraph{Rendering Results:}
Table~\ref{tab:experimental_results} reports the quantitative results for surgical scene reconstruction. On the EndoNeRF dataset (averaged over both pulling and cutting sequences), FE-4DGS outperforms all baselines on SSIM, PSNR and LPIPS, while handling semantic feature rendering in real-time. On the SCARED dataset, FE-4DGS still achieves the highest PSNR and lowest LPIPS while matching SSIM, demonstrating robustness afforded by our semantic integration. Figure~\ref{fig:render} shows example frames from the EndoNeRF cutting and pulling sets: the 1.3 dB PSNR gain over EndoGaussian corresponds to cleaner and more coherent textures in highly reflective regions. In the cutting example, there is also slightly improved rendering in the occluded regions which were missed in the EndoGaussian renders. Given these results, the incorporation of semantic features enhances geometric consistency during reconstruction. Additional renderings for EndoVis18 are found in Figure~\ref{fig:render2}.

\paragraph{Rendering Speed:}
A key advantage of 3DGS methods \citep{kerbl20233dgaussiansplattingrealtime} is the improved rendering speed relative to NeRF-based approaches \citep{mildenhall2020nerfrepresentingscenesneural}. 
Despite the additional complexity for semantic processing, FE-4DGS maintains competitive FPS and model size. On the NVIDIA RTX 5090 GPU, we achieve an average rendering speed of 287.95 FPS on the EndoNeRF dataset, at 392.5 MB. The tradeoff in speed is incurred because FE-4DGS performs real time rendering of RGB, depth, and semantic features in one pass, while previous works only render RGB and depth~\cite{liu2024endogaussianrealtimegaussiansplatting}. Despite the compromise in rendering speed, 287.95 FPS remains clinically substantial for typical endoscopic displays. Model size and FPS comparison can be found in Table~\ref{tab:model_sizes} and \ref{tab:fps_same_gpu}. We also note that FPS can be a hardware dependent metric and should not be overanalyzed.

\input{tables/model_sizes}
\input{tables/FPS}



\paragraph{Results on Segmentation:}
Tables~\ref{tab:seg_bin} and \ref{tab:seg_multi} report the binary and multi-label segmentation performance on the EndoVis18 dataset. For multi-label segmentation, Table~\ref{tab:seg_multi} shows the averaged results. We show the detailed per-class performance in Table~\ref{tab:avg_results} of the appendix. In the binary setting, FE-4DGS with \(F_{\text{feat}}\) achieves the best IoU, DSC, and recall scores, while maintaining competitive precision compared to state-of-the-art models such as SAM 2 \citep{ravi2024sam2segmentimages}. For multi-label segmentation, FE-4DGS outperforms SAM (ViT-B/H) and MedSAM across almost all metrics, though SAM 2 shows superior performance in some classes. This discrepancy suggests that while FE-4DGS captures high-level semantic features effectively, further refinement is needed to encode more class-specific details.
Yet FE-4DGS still delivers superior performance in binary segmentation and maintains strong performance in multi-label segmentation, outperforming its teacher model (i.e., ViT-H). This showcases the advantage of directly incorporating semantic features during training as opposed to simple post-hoc application of 2D segmentation models. Qualitative results are provided in Figure~\ref{fig:qual}.

\input{tables/feature_dim}

\subsection{Ablation Study}

\paragraph{Component Analysis:} Table~\ref{tab:feg-comp-full} presents an ablation study highlighting the contribution of each key component in FE-4DGS to the overall rendering quality. Removing the semantic deformation decoder \(F_{\text{feat}}\) results in degradation across all metrics, most notably in PSNR and \(L_1\) feature loss, indicating that the absence of fine-grained semantic updates hinders the model's ability to capture detailed semantic variations. Thus, leveraging physical deformation cues encoded in extracted features, \(h_\mu\), \(h_{\mathrm{R}}\), \(h_{\mathrm{S}}\), \(h_{\mathrm{o}}\), enables coherent, physically grounded updates for semantic features, leading to improved reconstructed feature maps and better downstream segmentation performance. Similarly, omitting the per-pixel \(L_1\) feature loss significantly degrades the alignment between the rendered and ground truth semantic features from the foundation model (SAM), as reflected by the increased \(L_1\) error. This misalignment confirms the importance of explicit semantic supervision. Lastly, excluding the HexPlane \citep{cao2023hexplanefastrepresentationdynamic} leads to a drastic decline in reconstruction quality, underscoring its importance in providing deformation representations.

\paragraph{Feature Dimension Sweep:}
Table~\ref{tab:ablate_featdim} varies the distilled feature dimensionality from 16 to 256. We observe a monotonic improvement in reconstruction and perceptual quality up to 128 (best PSNR/SSIM and lowest LPIPS), alongside the lowest per-pixel \(L_1\) feature error at 128. Increasing capacity further to 256 yields marginal regression in PSNR/SSIM and a higher feature \(L_1\), suggesting mild over-parameterization under the same training budget. These results indicate that 128 provides the best accuracy–capacity trade-off for reliable semantic field distillation and downstream tasks.

\input{tables/hexplane}
\paragraph{HexPlane Encoder Capacity:}
Table~\ref{tab:ablate_hexplane} compares HexPlane encoder widths of 32 and 64 while holding all other settings fixed. Doubling width from 32 to 64 consistently improves PSNR/SSIM and reduces LPIPS and the feature \(L_1\) loss. In FE-4DGS, the encoder must represent both standard Gaussian attributes and the additional semantic feature property; the larger width reduces underfitting in this joint representation space without incurring prohibitive runtime or memory costs. We therefore adopt width = 64 as the default.

\section{Conclusion}
\label{sec:conclusion}
We introduced FeatureEndo-4DGS (FE-4DGS), the first system to achieve real-time surgical scene reconstruction and multi-label semantic segmentation of dynamic endoscopic videos. By distilling features from 2D segmentation foundation models into the 4D rendering process, FE-4DGS consistently improves reconstruction quality over existing baselines while maintaining comparable FPS and model size. On binary and multi-label benchmarks, it matches or surpasses state-of-the-art 2D models, performing segmentation directly during rendering.

Our main limitation is the lack of high-quality endoscopic datasets with both reconstruction and segmentation annotations. For example, EndoVis18 \citep{allan20202018roboticscenesegmentation} provides masks but suffers from low image quality that hinders rendering. We hope this work motivates the collection of richer multimodal data—including text, images, and audio—that can be seamlessly integrated into our pipeline. Beyond reconstruction and segmentation, FE-4DGS also opens paths toward language-guided editing and promptable segmentation, as explored in Feature 3DGS \citep{zhou2024feature3dgssupercharging3d} and NeRF- \citep{kobayashi2022decomposingnerfeditingfeature}.

\acks{We thank Haohong Lin for the valuable discussions.}

\bibliography{jmlr-sample}

\appendix

\section{Feature-Spatiotemporal (FST) Deformation Module}
\label{apd:fst}
This supplementary material provides additional implementation and dataset details not included in the main paper.

\begin{figure*}[!ht]
    \centering
    \includegraphics[width=0.8\linewidth]{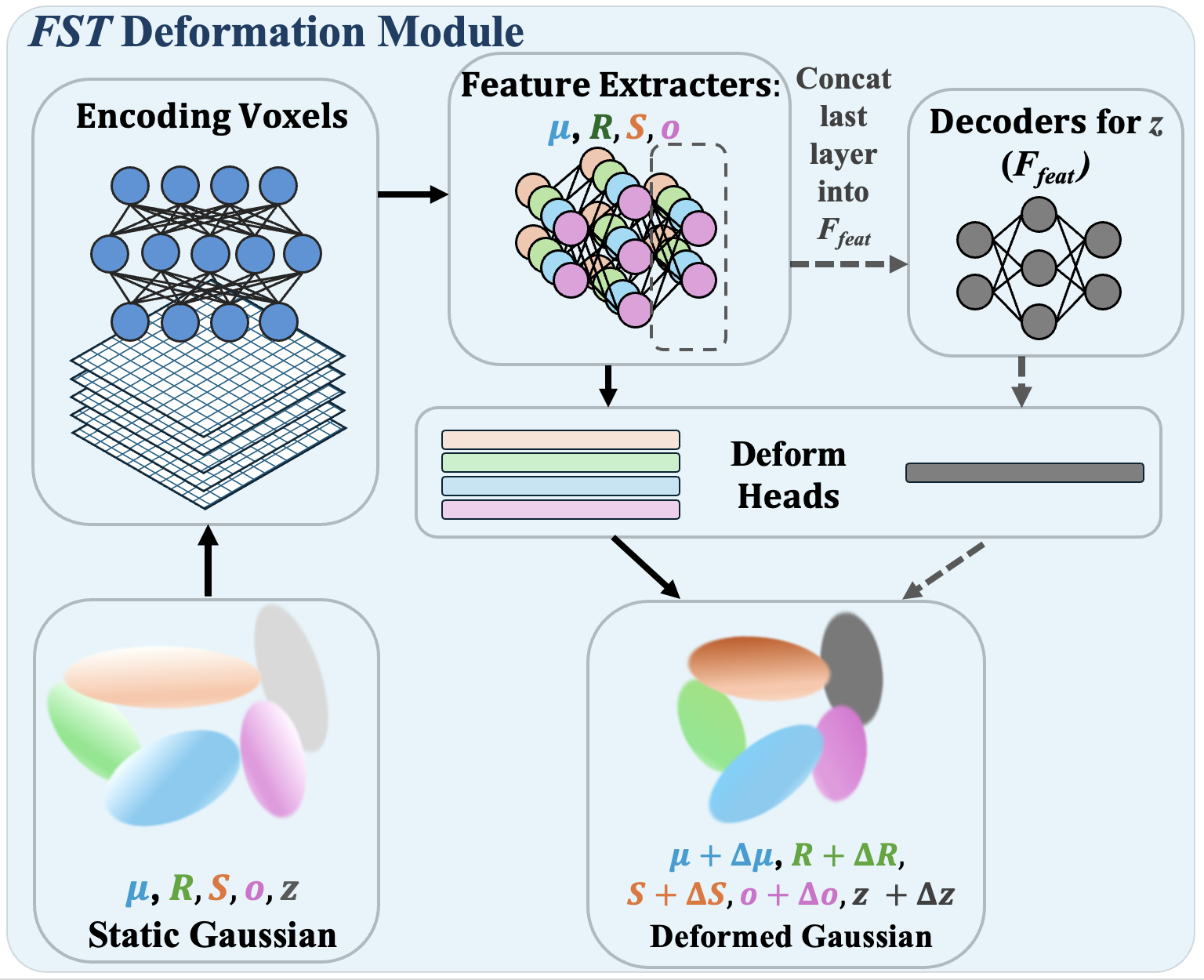}
    \caption{Overview of the FST feature deformation module found in FE-4DGS, it includes the HexPlanes \citep{cao2023hexplanefastrepresentationdynamic} and requires decoder architectures for updating position, rotation, scale, opacity, and semantic features of Gaussians.}
    \label{fig:feg_sup}
\end{figure*}

The FST deformation module is detailed in Figure~\ref{fig:feg_sup}, which provides a visual supplement of Section~\ref{sec:fst}. The implementation details for FST are as follows: for the HexPlane encoder, we follow the same configuration as given in EndoGaussian \cite{liu2024endogaussianrealtimegaussiansplatting}. That is, grid dimension of size 2, input coordinate dimension of size 4, output coordinate dimension of size 32, and resolutions of [64,64,64,100]. For the HexPlane decoder MLP, we set the model width W to 64 and the depth D to 8. Each feature extractor follows the same structure consisting of 2 linear layers that downsample the embeddings to W//2 before passing into deformation heads. The semantic feature MLP is structured similarly, with 2 linear layers where the second layer projects the embedding back into the feature size of 128 corresponding to the number of semantic channels we use in practice. 

\begin{figure*}[!htb]
    \centering
    \includegraphics[width=1\linewidth]{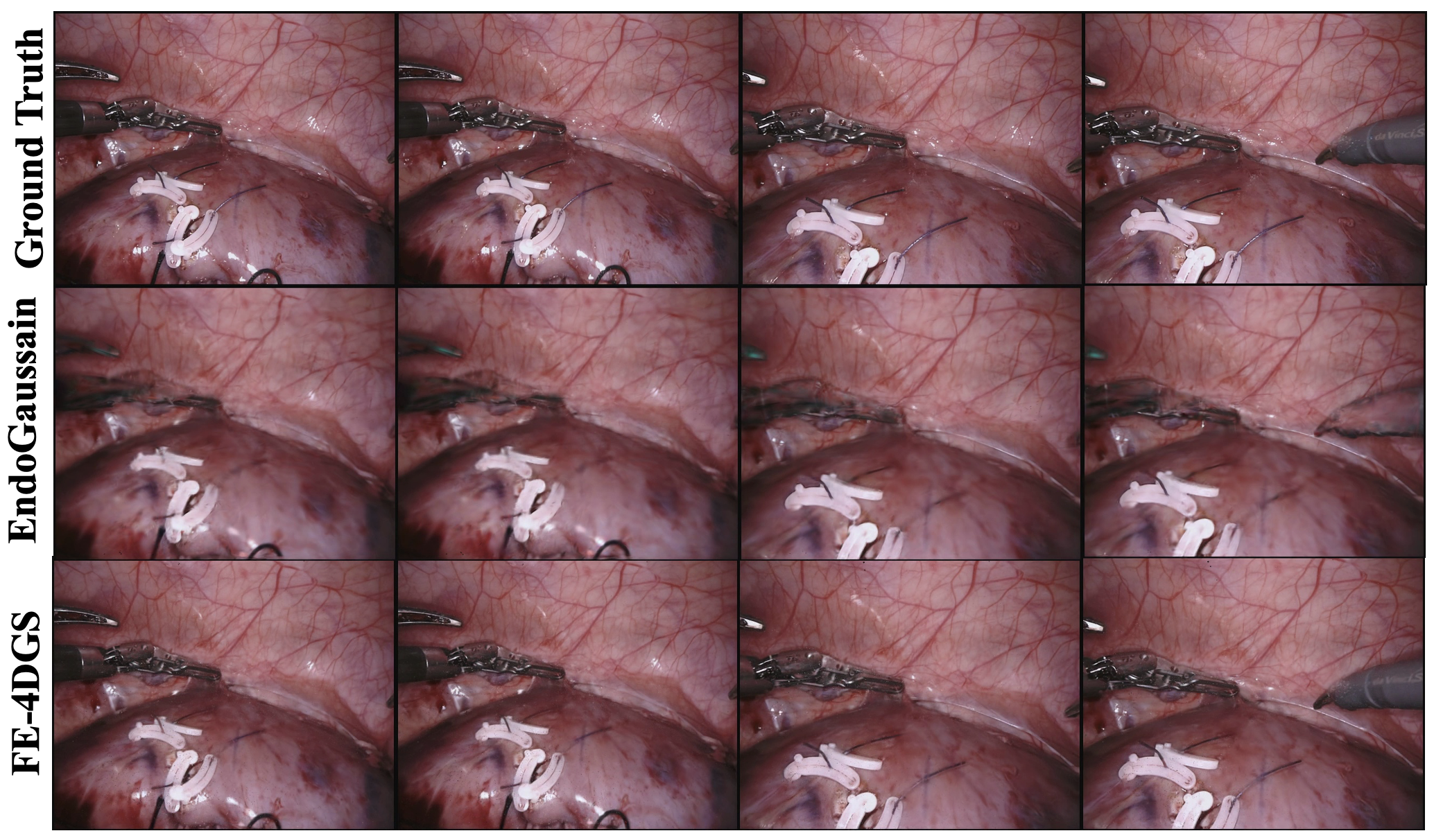}
    \caption{Comparison of qualitative renderings between EndoGaussian \citep{liu2024endogaussianrealtimegaussiansplatting}, FE-4DGS, and ground truth on the EndoVis18 dataset \citep{allan20202018roboticscenesegmentation}. We can see that across all scenes, FE-4DGS and EndoGaussian experience similar performances, although FE-4DGS is able to capture semantic features which can later be used for segmentation.}
    \label{fig:render2}
\end{figure*}

\section{Dataset Details}
\label{sec:app_a}

EndoNeRF \citep{wang2022neuralrenderingstereo3d} comprises 2 in-vivo prostatectomy cases captured with stereo cameras from a single viewpoint. The dataset features non-rigid deformations and tool occlusions. We also use the Stereo Correspondence and Reconstruction of Endoscopic Data (SCARED) \citep{allan2021stereocorrespondencereconstructionendoscopic} dataset from the 2022 Endoscopic Vision Challenge, which consists of RGB-D scans of porcine cadaver abdominal anatomy collected using a da Vinci Xi endoscope and a projector.

Within the EndoNeRF dataset, the pulling video contains 63 frames. We select every eighth frame for testing, resulting in 58 training frames and 5 test frames. The cutting video comprises 156 frames and is split in a similar manner, yielding 136 training frames and 20 test frames.

For segmentation, we employ the Robotic Scene Segmentation Sub-Challenge from the 2018 Endoscopic Vision Challenge (EndoVis18) \citep{allan20202018roboticscenesegmentation}. This dataset contains 4 sequences, each with 149 frames depicting diverse surgical scenes with annotations for various anatomical and tool components. In total, there are 10 labels in this multi-label segmentation dataset: background tissue, instrument shaft, instrument clasper, instrument wrist, kidney parenchyma, covered kidney, thread, clamps, suturing needles, suction instrument, and small intestine. The original images are sized at 1280\,$\times$\,1024 pixels and are downsampled to 640\,$\times$\,512 pixels for training. Although the original EndoVis18 videos were captured at 60\,Hz, they were subsampled to 2\,Hz to reduce labeling efforts. Furthermore, sequences with minimal movement were removed, resulting in the final 149-frame video. These postprocessing steps present significant challenges for high-fidelity surgical scene reconstruction. To prepare these images for FE-4DGS, we first obtain stereo depth masks using the lite stereo transformer \citep{li2021revisitingstereodepthestimation}. Additionally, we prepare a camera poses and intrinsics file in the LLFF (Local Light Field Fusion) format \citep{mildenhall2019locallightfieldfusion}. As a simplification, we assume a single viewpoint and set all camera poses to the identity to avoid interference from ill-calibrated poses.

To demonstrate the utility of FE-4DGS for segmentation, we select 60 frames from sequence 1, which depict a continuous surgical scene of the kidney. In these frames, a clip applier tool is used, and parts of the small intestine appear in the foreground. These frames were chosen for training because the scene contains multiple key segmentation targets, including the kidney, small intestine, clamps, and all components of grasping instruments (i.e., instrument head, instrument wrist, and instrument clasper). These targets vary in size, segmentation complexity, and instance count per frame, presenting a challenging multi-label segmentation task suitable for evaluating our model. Due to substantial tissue and tool movements and a camera repositioning midway through the scene, we split these 60 frames into 2 segments for rendering. The first segment, comprising 33 frames, is processed through our FE-4DGS pipeline, while the latter 27 frames are rendered as a separate scene. Aside from the mid-scene movements, the camera remains fairly stable, with smooth transitions and consistent overlap in scene coverage, ensuring a well-constrained reconstruction. During segmentation, we aggregate the feature maps from these 2 segments to benchmark against standalone SAM models.

\section{Additional Details of Integrating Semantic Features into FE-4DGS}
\label{sec:app_b}
This supplementary material provides additional details on integrating semantic features into FE-4DGS that were not included in the main paper.

For segmentation experiments, we use SAM's pretrained ViT-H encoder to generate a ground truth feature map for each frame in our video snippet. As detailed in SAM \citep{kirillov2023segment}, the image encoder is pretrained via MAE and employs 14×14 windowed attention along with 4 equally spaced global attention blocks. The resulting 256 feature maps have dimensions of 64×64, which are then resized to 51×64 to match the aspect ratio of the training image. This resizing is accomplished by simply cropping the feature map height, as done in Feature 3DGS \citep{zhou2024feature3dgssupercharging3d}. Since only the 51×64 portion of the 64×64 feature map contains meaningful semantic information and the rest is padding, we crop the side corresponding to the longer dimension of the original image.

For segmentation, we use the same configuration as with the EndoNeRF cutting dataset. We note that no hyperparameter tuning has been performed for the EndoVis dataset, and performance metrics may be further improved with additional tuning. After running FE-4DGS, the generated feature maps are directly fed into the pretrained SAM decoder to produce multi-label segmentations. Prior to this step, all feature maps are restored to their original aspect ratio of 64×64 via bilinear interpolation. For the segmentation prompt, bounding boxes computed from the ground truth mask for each class are used. If a class appears in multiple masks within an image, bounding boxes are computed for each instance.

\section{Fine-tuning SAM Details}
\label{sec:app_c}

We use the SAM model with the ViT-B encoder and standard pretrained weights. Instead of fully fine-tuning all layers, we adopt an "adapter" training strategy \citep{Gu_2025}, fine-tuning only the adapter layer parameters while keeping the rest of the SAM backbone frozen. This significantly reduces VRAM consumption during training, allowing us to work within our limited computational resources. For training, we use sequences 2, 3, and 4 from EndoVis18, each containing 149 frames, for a total of 447 images. Sequence 1, which also contains 149 images, is reserved for validation and is excluded from training because it is later used for zero-shot segmentation in Tables~\ref{tab:seg_bin} and~\ref{tab:seg_multi}.

The preprocessing steps are straightforward: images are normalized and resized. The loss function is defined as the sum of the dice loss and the cross entropy. We employ an Adam optimizer with a base learning rate of \(1 \times 10^{-3}\) and a StepLR scheduler with a step size of 10 and \(\gamma = 0.5\). The batch size is set to 4, and validation is performed every other epoch. Hyperparameters are not tuned; training continues until the validation loss does not decrease for more than 20 epochs. In total, we train for 66 epochs, achieving the best validation score of 0.955 at epoch 46.

Subsequently, we extract embeddings from this fine-tuned model to serve as the ground truth feature embeddings for our FE-4DGS pipeline. Zero-shot segmentation is performed using the semantic feature maps generated by FE-4DGS on 60 selected image frames from EndoVis18 for binary segmentation. For this experiment, we follow the same 7:1 train-test split as in EndoNeRF, which enables us to obtain performance metrics on the test set. Segmentation performance is measured via micro-averaging across all classes on the train image frames, with each label weighted by its mask size, and we observe a marked improvement when using these fine-tuned features compared to the baseline.

Table~\ref{tab:finetune-single} compares segmentation and rendering performance using the original versus the fine-tuned SAM model (trained on EndoVis18 sets 2, 3, and 4). Fine-tuning significantly improves segmentation metrics—IoU and DSC—demonstrating that adapting SAM to the domain-specific characteristics of endoscopic scenes yields more accurate semantic cues. These enhanced semantic features, in turn, facilitate better guidance during rendering, as reflected by SSIM and PSNR scores. Overall, the results affirm that task-specific refinement of the segmentation backbone is important in boosting both segmentation fidelity and the subsequent rendering quality.

\input{tables/ft_sam}

\section{Additional Results}
\input{tables/table_seg_avg_class}
\subsection{Class-wise Multi-label Segmentation Results}
We provide additional results on each individual class present in the EndoVis18 dataset in Table~\ref{tab:avg_results}.
\input{tables/hp}
\input{tables/teacher_model}
\subsection{Teacher Model Choice}
Table~\ref{tab:ablate_teacher} evaluates various segmentation teacher models for semantic supervision. We find that all models achieve comparable performance, with SAM ViT-H showing a slightly higher PSNR than the others. Therefore, we primarily use SAM ViT-H as the teacher model.

\section{Additional Visualizations}


\subsection{Additional Rendering Visualizations}
We provide additional visualizations of rendering on the EndoVis18 dataset in Figure~\ref{fig:render2}.

\section{Training Hyperparameters}
\label{apd:hp}

In Table~\ref{tab:hyperparams}, we present the hyperparameter configurations used in our experiments for the EndoNeRF and SCARED datasets. Most other hyperparameters remain fixed at the default values provided in the code.

We observed stability issues during the fine stage on the EndoNeRF dataset, likely due to our model's enhanced rendering performance during the coarse stage, which is attributed to our revamped rasterizer. An excessively high PSNR during the coarse phase—reaching up to 50—indicated overfitting during initialization with a single image, thereby hindering subsequent deformation optimization. Although the first 100–300 iterations of the fine stage showed promising convergence, the model occasionally experienced an abrupt drop in PSNR, sometimes as low as 5. To maintain stable training dynamics and ensure generalizable deformation capabilities, it was necessary to adjust hyperparameters, particularly by reducing the learning rates and fine-tuning gradient propagation strategies. Furthermore, the coarse initialization stage must be capped to conclude at a lower PSNR to ensure stability during the fine deformation stage.

\end{document}

%% file: tables/main_1.tex
\definecolor{LightGreen}{RGB}{204, 255, 204} 

\begin{table*}[t]
\small
\centering
\caption{Baselines for surgical scene reconstruction on ENDONERF~\citep{wang2022neuralrenderingstereo3d} and SCARED~\citep{allan2021stereocorrespondencereconstructionendoscopic}.}
\label{tab:experimental_results}
\resizebox{0.95\textwidth}{!}{%
\begin{tabular}{l l c c c}
\toprule
\textbf{Dataset} & \textbf{Method} & \textbf{LPIPS (↓)} & \textbf{SSIM (↑)} & \textbf{PSNR (↑)}  \\
\midrule
\multirow{8}{*}{ENDONERF~\cite{wang2022neuralrenderingstereo3d}} 
& EndoNeRF~\cite{wang2022neuralrenderingstereo3d} & 0.09 & 0.93 & 36.06  \\
& EndoSurf~\cite{zha2023endosurfneuralsurfacereconstruction} & 0.07 & 0.95 & 36.53  \\
& LerPlane-9k~\cite{yang2023neurallerplanerepresentationsfast} & 0.08 & 0.93 & 34.99  \\
& LerPlane-32k~\cite{yang2023neurallerplanerepresentationsfast} & 0.05 & 0.95 & 37.38  \\
& Endo-4DGS~\cite{huang2024endo} & 0.04 & 0.96 & 37.21 \\
& EndoGS~\cite{Zhou2021EMDQSLAMRH}  & 0.05 & 0.96 & 37.29  \\
& EndoGaussian~\cite{liu2024endogaussianrealtimegaussiansplatting} & 0.05 & 0.96 & 37.78  \\
& LGS~\cite{10.1007/978-3-031-72384-1_62} & 0.07 & 0.96 & 37.48  \\
  & {\bfseries FE-4DGS (Ours)} 
    & {\bfseries 0.03} 
    & {\bfseries 0.97} 
    & {\bfseries 39.08}  \\
\midrule
\multirow{5}{*}{SCARED~\cite{allan2021stereocorrespondencereconstructionendoscopic}} 
& EndoNeRF~\cite{wang2022neuralrenderingstereo3d} & 0.40 & 0.77 & 24.35  \\
& EndoSurf~\cite{zha2023endosurfneuralsurfacereconstruction} & 0.36 & 0.80 & 25.02  \\
& EndoGaussian~\cite{liu2024endogaussianrealtimegaussiansplatting} & 0.27 & \textbf{0.83} & 26.89  \\
& LGS~\cite{10.1007/978-3-031-72384-1_62} & 0.30 & \textbf{0.83} & 27.05  \\
  & {\bfseries FE-4DGS (Ours)} 
    & {\bfseries 0.23} 
    & {\bfseries 0.83} 
    & {\bfseries 27.28}  \\
\bottomrule
\end{tabular}%
}
\end{table*}


%% file: tables/main_2.tex
\begin{table*}[t]
\centering
\small
\caption{Comparison across different segmentation foundation models and two modes of FE-4DGS for binary segmentation on the EndoVis18 dataset \citep{allan2021stereocorrespondencereconstructionendoscopic}. FE-4DGS (w/o \(F_{\mathrm{feat}}\)) denotes the removal of the \(F_{\mathrm{feat}}\) network, where semantic features do not receive additional temporal updates from the FST Deformation Module (i.e. semantic features only supervised by L1-feature loss). 
Values are reported as the mean $\pm$ half-width of the 95\% bootstrap confidence interval (B = 10{,}000).}
\label{tab:seg_bin}
\resizebox{0.75\textwidth}{!}{%
\begin{tabular}{l c c c c}
\toprule
\textbf{Model} & \textbf{IoU (↑)} & \textbf{DSC (↑)} & \textbf{Recall (↑)} & \textbf{Precision (↑)} \\
\midrule
SAM ViT-B & $0.85 \pm 0.02$ & $0.92 \pm 0.01$ & $0.95 \pm 0.01$ & $0.89 \pm 0.02$ \\
SAM ViT-H & $0.86 \pm 0.02$ & $0.92 \pm 0.01$ & $0.94 \pm 0.01$ & $0.91 \pm 0.02$ \\
SAM 2 (Hiera-T) & $0.82 \pm 0.03$ & $0.90 \pm 0.02$ & $0.89 \pm 0.01$ & $0.91 \pm 0.02$ \\
SAM 2 (Hiera-H) & $0.84 \pm 0.02$ & $0.92 \pm 0.01$ & $0.93 \pm 0.01$ & \textbf{$0.93 \pm 0.02$} \\
MedSAM ViT-B & $0.84 \pm 0.02$ & $0.91 \pm 0.01$ & $0.92 \pm 0.01$ & $0.91 \pm 0.02$ \\
FE-4DGS (w/o \(F_{\mathrm{feat}}\)) & $0.79 \pm 0.03$ & $0.88 \pm 0.02$ & $0.86 \pm 0.01$ & $0.90 \pm 0.03$ \\
\textbf{FE-4DGS (Ours)} &
\textbf{$0.88 \pm 0.03$} &
\textbf{$0.93 \pm 0.02$} &
\textbf{$0.96 \pm 0.01$} &
{$0.91 \pm 0.03$} \\
\bottomrule
\end{tabular}%
}
\end{table*}

%% file: tables/main_3.tex
\definecolor{LightGreen}{RGB}{204, 255, 204} 

\begin{table*}[t]
\centering
\small
\caption{Weight-averaged scores for multi-label segmentation on the EndoVis18 dataset \citep{allan2021stereocorrespondencereconstructionendoscopic}.}
\label{tab:seg_multi}
\resizebox{0.75\textwidth}{!}{%
\begin{tabular}{l c c c c}
\toprule
\textbf{Model} & \textbf{IoU (↑)} & \textbf{DSC (↑)} & \textbf{Recall (↑)} & \textbf{Precision (↑)} \\
\midrule
SAM ViT-B & $0.67 \pm 0.03$ & $0.73 \pm 0.03$ & $0.75 \pm 0.02$ & $0.79 \pm 0.03$ \\
SAM ViT-H & $0.68 \pm 0.02$ & $0.76 \pm 0.02$ & $0.74 \pm 0.02$ & $0.84 \pm 0.01$ \\
SAM 2 (Hiera T) & $0.77 \pm 0.02$ & $0.85 \pm 0.01$ & $0.88 \pm 0.01$ & $0.84 \pm 0.01$ \\
SAM 2 (Hiera H) & {\bfseries\boldmath $0.79 \pm 0.02$} & {\bfseries\boldmath $0.87 \pm 0.01$} & {\bfseries\boldmath $0.87 \pm 0.01$} & {\bfseries\boldmath $0.88 \pm 0.01$} \\
MedSAM ViT-B & $0.66 \pm 0.03$ & $0.75 \pm 0.02$ & $0.86 \pm 0.01$ & $0.72 \pm 0.02$ \\
FE-4DGS (w/o \(F_{\mathrm{feat}}\)) & $0.63 \pm 0.02$ & $0.73 \pm 0.01$ & $0.72 \pm 0.02$ & $0.82 \pm 0.01$ \\
\textbf{FE-4DGS (Ours)} & $0.70 \pm 0.01$ & $0.77 \pm 0.01$ & $0.77 \pm 0.01$ & $0.84 \pm 0.01$ \\
\bottomrule
\end{tabular}%
}
\end{table*}

%% file: tables/combined2.tex
\definecolor{LightGreen}{RGB}{204,255,204} 

\begin{table*}[t]
\centering
\small
\caption{Ablation on removing different components of FE-4DGS.}
\label{tab:feg-comp-full}
\resizebox{0.75\textwidth}{!}{%
\begin{tabular}{l c c c c}
\toprule
\textbf{Model}  & \textbf{SSIM (↑)} & \textbf{PSNR (↑)} & \textbf{LPIPS (↓)} & \textbf{L1 Feature Loss (↓)}\\
\midrule
w/o $F_{\text{feat}}$  &  0.97 & 38.56 & 0.04 & 0.10 \\
w/o L1 Feature Loss  & 0.97  & 38.43 & 0.04 & 0.18 \\
w/o HexPlane  &  0.88 & 26.65 & 0.17 & 0.06 \\
\textbf{FE-4DGS (Ours)} & \textbf{0.97} & \textbf{39.08} & \textbf{0.03} & \textbf{0.03} \\
\bottomrule
\end{tabular}%
}
\end{table*}


%% file: tables/model_sizes.tex
\begin{table}[t]
\centering
\small
\caption{Comparison of baseline model sizes and FE-4DGS in megabytes (MB).}
\label{tab:model_sizes}
\resizebox{\columnwidth}{!}{%
\begin{tabular}{lccc}
\toprule
\textbf{Model} & \textbf{FE-4DGS} & \textbf{EndoGaussian} & \textbf{EndoNeRF} \\
\midrule
\textbf{Model Size (MB) (↓)} & 392.50 & 334.50 & \textbf{13.00} \\
\bottomrule
\end{tabular}%
}
\end{table}

%% file: tables/FPS.tex
\begin{table}[t]
\centering
\small
\caption{FPS comparison across baselines on the same NVIDIA GPU.}
\label{tab:fps_same_gpu}
\resizebox{\columnwidth}{!}{%
\begin{tabular}{lccc}
\toprule
\textbf{GPU} & \textbf{FE-4DGS} & \textbf{EndoGaussian} & \textbf{EndoNeRF} \\
\midrule
\textbf{RTX A6000 FPS (↑)} & 61.32 & \textbf{140.36} & 0.02 \\
\textbf{RTX 5090 FPS (↑)} & 287.95 & \textbf{399.32} & 0.36 \\
\bottomrule
\end{tabular}%
}
\end{table}

%% file: tables/feature_dim.tex
\begin{table}[t]
\captionsetup{width=\columnwidth}
\centering
\small
\caption{Reconstruction results on varying the dimensionality of the distilled semantic features.}
\label{tab:ablate_featdim}
\resizebox{\columnwidth}{!}{%
\begin{tabular}{lcccc}
\toprule
\textbf{Dim} & \textbf{PSNR (↑)} & \textbf{SSIM (↑)} & \textbf{LPIPS (↓)} & \textbf{L1 feature Loss (↓)} \\
\midrule
16 & 38.52 & 0.96 & 0.04 & 0.07 \\
32 & 38.81 & 0.97 & 0.03 & 0.05 \\
64 & 38.98 & 0.97 & 0.03 & 0.04 \\
\textbf{128} & \textbf{39.08} & \textbf{0.97} & \textbf{0.03} & \textbf{0.03} \\
256 & 38.99 & 0.97 & 0.03 & 0.03 \\
\bottomrule
\end{tabular}%
}
\end{table}

%% file: tables/hexplane.tex
\begin{table}[t]
\centering
\small
\caption{Ablation study on various net widths of HexPlane.}
\label{tab:ablate_hexplane}
\resizebox{\columnwidth}{!}{%
\begin{tabular}{lcccc}
\toprule
\textbf{Net Width} & \textbf{PSNR (↑)} & \textbf{SSIM (↑)} & \textbf{LPIPS (↓)} & \textbf{L1 feature Loss (↓)} \\
\midrule
32 & 38.63 & 0.96 & 0.03 & 0.04 \\
\textbf{64} & \textbf{39.08} & \textbf{0.97} & \textbf{0.03} & \textbf{0.03} \\
\bottomrule
\end{tabular}%
}
\end{table}

%% file: tables/ft_sam.tex
\begin{table}[t]
\centering
\caption{Fine-tuning SAM (EndoVis18 \citep{allan20202018roboticscenesegmentation}).}
\label{tab:finetune-single}
\resizebox{\columnwidth}{!}{%
\begin{tabular}{l c c c c}
\toprule
\textbf{Finetuned} & \textbf{IoU (↑)} & \textbf{DSC (↑)} & \textbf{SSIM (↑)} & \textbf{PSNR (↑)} \\
\midrule
X           & 0.60 & 0.73 & 0.67 & 20.43 \\
\checkmark  & \textbf{0.69} & \textbf{0.81} & \textbf{0.67} & \textbf{20.77} \\
\bottomrule
\end{tabular}%
}
\end{table}

%% file: tables/table_seg_avg_class.tex
\begin{table}[ht]\centering
\caption{Class-wise multi-label segmentation results on the EndoVis18 dataset.}
\label{tab:avg_results}
\resizebox{\columnwidth}{!}{%
\begin{tabular}{lrrrrr}
\toprule
\textbf{Model} & \textbf{Class} & \textbf{IoU (↑)} & \textbf{DSC (↑)} & \textbf{Recall (↑)} & \textbf{Precision (↑)} \\
\midrule
\multirow{6}{*}{FE-4DGS (ViT-H)} 
 & kidney             & 0.87 & 0.93 & 0.96 & 0.90 \\
 & small intestine    & 0.83 & 0.90 & 0.91 & 0.90 \\
 & instrument shaft   & 0.34 & 0.38 & 0.36 & 0.67 \\
 & instrument clasper & 0.26 & 0.36 & 0.32 & 0.66 \\
 & clamps             & 0.27 & 0.40 & 0.31 & 0.70 \\
 & instrument wrist   & 0.69 & 0.81 & 0.91 & 0.75 \\
\midrule
\multirow{6}{*}{SAM (ViT-H)} 
 & kidney             & 0.85 & 0.91 & 0.92 & 0.92 \\
 & small intestine    & 0.85 & 0.92 & 0.93 & 0.90 \\
 & instrument shaft   & 0.32 & 0.37 & 0.34 & 0.66 \\
 & instrument clasper & 0.26 & 0.36 & 0.31 & 0.65 \\
 & clamps             & 0.28 & 0.40 & 0.32 & 0.69 \\
 & instrument wrist   & 0.76 & 0.85 & 0.87 & 0.85 \\
\midrule
\multirow{6}{*}{SAM (ViT-B)} 
 & kidney             & 0.80 & 0.89 & 0.93 & 0.86 \\
 & small intestine    & 0.85 & 0.92 & 0.93 & 0.90 \\
 & instrument shaft   & 0.32 & 0.37 & 0.34 & 0.60 \\
 & instrument clasper & 0.26 & 0.36 & 0.32 & 0.63 \\
 & clamps             & 0.28 & 0.40 & 0.32 & 0.69 \\
 & instrument wrist   & 0.76 & 0.85 & 0.86 & 0.85 \\
\midrule
\multirow{6}{*}{MedSAM (ViT-B)} 
 & kidney             & 0.83 & 0.90 & 0.90 & 0.91 \\
 & small intestine    & 0.84 & 0.91 & 0.94 & 0.88 \\
 & instrument shaft   & 0.32 & 0.36 & 0.34 & 0.74 \\
 & instrument clasper & 0.25 & 0.35 & 0.30 & 0.74 \\
 & clamps             & 0.24 & 0.36 & 0.28 & 0.73 \\
 & instrument wrist   & 0.71 & 0.81 & 0.79 & 0.89 \\
\midrule
\multirow{6}{*}{SAM 2 (Hiera-T)} 
 & kidney             & 0.85 & 0.92 & 0.90 & 0.94 \\
 & small intestine    & 0.83 & 0.88 & 0.87 & 0.91 \\
 & instrument shaft   & 0.84 & 0.90 & 0.93 & 0.90 \\
 & instrument clasper & 0.36 & 0.48 & 0.59 & 0.58 \\
 & instrument wrist   & 0.40 & 0.46 & 0.49 & 0.45 \\
 & clamps             & 0.66 & 0.79 & 0.88 & 0.72 \\
\midrule
\multirow{6}{*}{SAM 2 (Hiera-H)} 
 & kidney             & 0.87 & 0.93 & 0.92 & 0.94 \\
 & small intestine    & 0.90 & 0.94 & 0.95 & 0.94 \\
 & instrument shaft   & 0.83 & 0.90 & 0.92 & 0.90 \\
 & instrument clasper & 0.69 & 0.67 & 0.84 & 0.69 \\
 & instrument wrist   & 0.38 & 0.45 & 0.50 & 0.43 \\
 & clamps             & 0.66 & 0.79 & 0.87 & 0.72 \\
\bottomrule
\end{tabular}}
\end{table}

%% file: tables/hp.tex


\begin{table*}[ht!]
\centering
\caption{Hyperparameter Settings for pulling and cutting sets from EndoNeRF \citep{wang2022neuralrenderingstereo3d} and the SCARED dataset \citep{allan2021stereocorrespondencereconstructionendoscopic}.}
\label{tab:hyperparams}
\resizebox{\textwidth}{!}{%
\begin{tabular}{l c c c}
\toprule
\textbf{Hyperparameter} & \textbf{EndoNeRF Pulling} & \textbf{EndoNeRF Cutting} & \textbf{SCARED} \\
\midrule
Initial Points & 90,000 & 90,000 & 30,000 \\
Grid LR (Initial / Final) & 0.0032 / 0.0000032 & 0.0016 / 0.0000016 & 0.0016 / 0.000016 \\
Deformation LR (Initial / Final) & 0.00016 / 1.6e-7 & 0.0004 / 4e-7 & 0.00008 / 0.0000008 \\
Position LR (Initial / Final) & 0.00016 / 0.0000016 & 0.00016 / 0.0000016 & 0.00016 / 0.0000016 \\
Iterations (Coarse / Fine) & 1000 / 6000 & 1000 / 6000 & 1000 / 3000 \\
Percent Dense & 0.01 & 0.01 & 0.01 \\
Opacity Reset Interval & 6000 & 6000 & 3000 \\
Prune Interval & 6000 & 6000 & 3000 \\
Position LR Max Steps & 7000 & 7000 & 3000 \\
Deformation LR Delay Multiplier & 0.01 & 0.01 & 0.01 \\
Grid Dimensions & 2 & 2 & 2 \\
Input Coordinate Dim & 4 & 4 & 4 \\
Output Coordinate Dim & 64 & 64 & 32 \\
Multiresolution Levels & [1,2,4,8] & [1, 2, 4, 8] & [1, 2, 4, 8] \\
\bottomrule
\end{tabular}}%
\label{tab:hyperparams_comparison}
\end{table*}

%% file: tables/teacher_model.tex
\begin{table}[t]
\captionsetup{width=\columnwidth}
\centering
\small
\caption{Reconstruction performances on various segmentation teacher models.}
\label{tab:ablate_teacher}
\resizebox{\columnwidth}{!}{%
\begin{tabular}{lcccc}
\toprule
\textbf{Teacher} & \textbf{PSNR (↑)} & \textbf{SSIM (↑)} & \textbf{LPIPS (↓)} & \textbf{L1-feature Loss (↓)} \\
\midrule
SAM ViT-B  & 38.97 & 0.97 & 0.03 & 0.03 \\
\textbf{SAM ViT-H} & \textbf{39.08} & \textbf{0.97} & \textbf{0.03} & \textbf{0.03} \\
MedSAM-B   & 39.04 & 0.97 & 0.03 & 0.03 \\
\bottomrule
\end{tabular}%
}
\end{table}